\documentclass[11pt,a4paper]{article}
\usepackage[hyperref]{acl2017}
\usepackage{times}
\usepackage{latexsym}
\usepackage{amsmath}
\usepackage{url}
\usepackage{amssymb}
\usepackage{graphicx}
\usepackage{bbold}
\usepackage{tabularx}
\usepackage{url}

\aclfinalcopy 
\def\aclpaperid{19} 

\title{Generative Encoder-Decoder Models for Task-Oriented Spoken Dialog Systems with Chatting Capability}

\author{Tiancheng Zhao, Allen Lu, Kyusong Lee and Maxine Eskenazi\\
  Language Technologies Institute \\
  Carnegie Mellon University \\
  Pittsburgh, Pennsylvania, USA \\
  {\tt \{tianchez,arlu,kyusongl,max+\}@cs.cmu.edu}
  }

\date{}

\begin{document}
\maketitle
\begin{abstract}
Generative encoder-decoder models offer great promise in developing domain-general dialog systems. However, they have mainly been applied to open-domain conversations. This paper presents a practical and novel framework for building task-oriented dialog systems based on encoder-decoder models. This framework enables encoder-decoder models to accomplish slot-value independent decision-making and interact with external databases. Moreover, this paper shows the flexibility of the proposed method by interleaving chatting capability with a slot-filling system for better out-of-domain recovery. The models were trained on both real-user data from a bus information system and human-human chat data. Results show that the proposed framework achieves good performance in both offline evaluation metrics and in task success rate with human users.
\end{abstract}

\section{Introduction}
\label{sec:intro}
Task-oriented spoken dialog systems have transformed human-computer interaction by enabling people interact with computers via spoken language~\cite{raux2005let,young2006using,bohus2003ravenclaw}. The task-oriented SDS is usually domain-specific. The system creators first map the user utterances into semantic frames that contain domain-specific slots and intents using spoken language understanding (SLU)~\cite{de2008spoken}. Then a set of domain-specific dialog state variables is tracked to retain the context information over turns~\cite{williams2013dialog}. Lastly, the dialog policy decides the next move from a list of dialog acts that covers the expected communicative functions from the system.

Although the above approach has been successfully applied to many practical systems, it has limited ability to generalize to out-of-domain (OOD) requests and to scale up to new domains. For example, even within in a simple domain, real users often make requests that are not included in the semantic specifications. Due to this, proper error handling strategies that guide users back to the in-domain conversation are crucial to dialog success~\cite{bohus2005error}. Past error handling strategies were limited to a set of predefined dialog acts, e.g. request repeat, clarification etc., which constrained the system's capability in keeping users engaged. Moreover, there has been an increased interest in extending task-oriented systems to multiple topics~\cite{lee2009example,gavsic2015distributed} and multiple skills, e.g. grouping heterogeneous types of dialogs into a single system~\cite{zhao2016dialport}. Both cases require the system to be flexible enough to extend to new slots and actions.

Our goal is to move towards a domain-general task-oriented SDS framework that is flexible enough to expand to new domains and skills by removing domain-specific assumptions on the dialog state and dialog acts~\cite{bordes2016learning}. To achieve this goal, the neural encoder-decoder model\cite{cho2014learning,sutskever2014sequence} is a suitable choice, since it has achieved promising results in modeling open-domain conversations~\cite{vinyals2015neural,sordoni2015neural}. It encodes the dialog history using deep neural networks and then generates the next system utterance word-by-word via recurrent neural networks (RNNs). Therefore, unlike the traditional SDS pipeline, the encoder-decoder model is theoretically only limited by its input/output vocabulary.

A na\``{i}ve implementation of an encoder-decoder-based task-oriented system would use RNNs to encode the raw dialog history and generate the next system utterance using a separate RNN decoder. However, while this implementation might achieve good performance in an offline evaluation of a closed dataset, it would certainly fail when used by humans. There are several reasons for this: 1) real users can mention new entities that do not appear in the training data, such as a new restaurant name. These entities are, however, essential in delivering the information that matches users' needs in a task-oriented system. 2) a task-oriented SDS obtains information from a knowledge base (KB) that is constantly updated (``today's'' weather will be different every day), so simply memorizing KB results that occurred in the training data would produce false information. Instead, an effective model should learn to query the KB constantly to get the most up-to-date information. 3) users may give OOD requests (e.g. say, ``how is your day'', to a slot-filling system), which must be handled gracefully in order to keep the conversation moving in the intended direction.

This paper proposes an effective encoder-decoder framework for building task-oriented SDSs. We propose \textit{entity indexing} to tackle the challenges of out-of-vocabulary (OOV) entities and to query the KB. Moreover, we show the extensibility of the proposed model by adding chatting capability to a task-oriented encoder-decoder SDS for better OOD recovery. This approach was assessed on the Let's Go Bus Information data from the 1st Dialog State Tracking Challenge~\cite{williams2013dialog}, and we report performance on both offline metrics and real human users. Results show that this model attains good performance for both of these metrics. 


\section{Related Work}
\label{sec:related}
Past research in developing domain-general dialog systems can be broadly divided into three branches. The first one focuses on learning domain-independent dialog state representation while still using hand-crafted dialog act system actions. Researchers proposed the idea of extracting slot-value independent statistics as the dialog state~\cite{wang2015learning,gavsic2015policy}, so that the dialog state representation can be shared across systems serving different knowledge sources. Another approach uses RNNs to automatically learn a distributed vector representation of the dialog state by accumulating the observations at each turn~\cite{williams2016end,zhao2016towards,dhingra2016end,williams2017hybrid}. The learned dialog state is then used by the dialog policy to select the next action. The second branch of research develops a domain-general action space for dialog policy. Prior work replaced the domain-specific dialog acts with domain-independent natural language semantic schema as the action space of dialog managers~\cite{eshghi2014domain}, e.g. Dynamic Syntax~\cite{kempson2000dynamic}. More recently, Wen, et al.~\shortcite{wen2016network} have shown the feasibility of using an RNN as the decoder to generate the system utterances word by word, and the dialog policy of the proposed model can be fine tuned using reinforcement learning~\cite{su2016continuously}. Furthermore, to deal with the challenge of developing end-to-end task-oriented dialog models that are able to interface with external KB, prior work has unified the special KB query actions via deep reinforcement learning~\cite{zhao2016towards} and soft attention over the database~\cite{dhingra2016end}. The third branch strives to solve both problems at the same time by building an end-to-end model that maps an observable dialog history directly to the word sequences of the system's response. By using an encoder-decoder model, it has been successfully applied to open-domain conversational models~\cite{serban2015building,li2015diversity,li2016deep,zhao2017learning}, as well as to task oriented systems~\cite{bordes2016learning,yang2016reference,eric2017copy}. In order to better predict the next correct system action, this branch has focused on investigating various neural network architectures to improve the machine's ability to reason over user input and model long-term dialog context.

This paper is closely related to the third branch, but differs in the following ways: 1) these models are slot-value independent by leveraging domain-general entity recognizer, which is more extensible to OOV entities, 2) these models emphasize the interactive nature of dialog and address out-of-domain handling by interleaving chatting in task-oriented conversations, 3) instead of testing on a synthetic dataset, this approach focuses on real world use by testing the system on human users via spoken interface.

\section{Proposed Method}
\label{sec:model}
\begin{figure*}[ht]
    \centering
    \includegraphics[width=16cm]{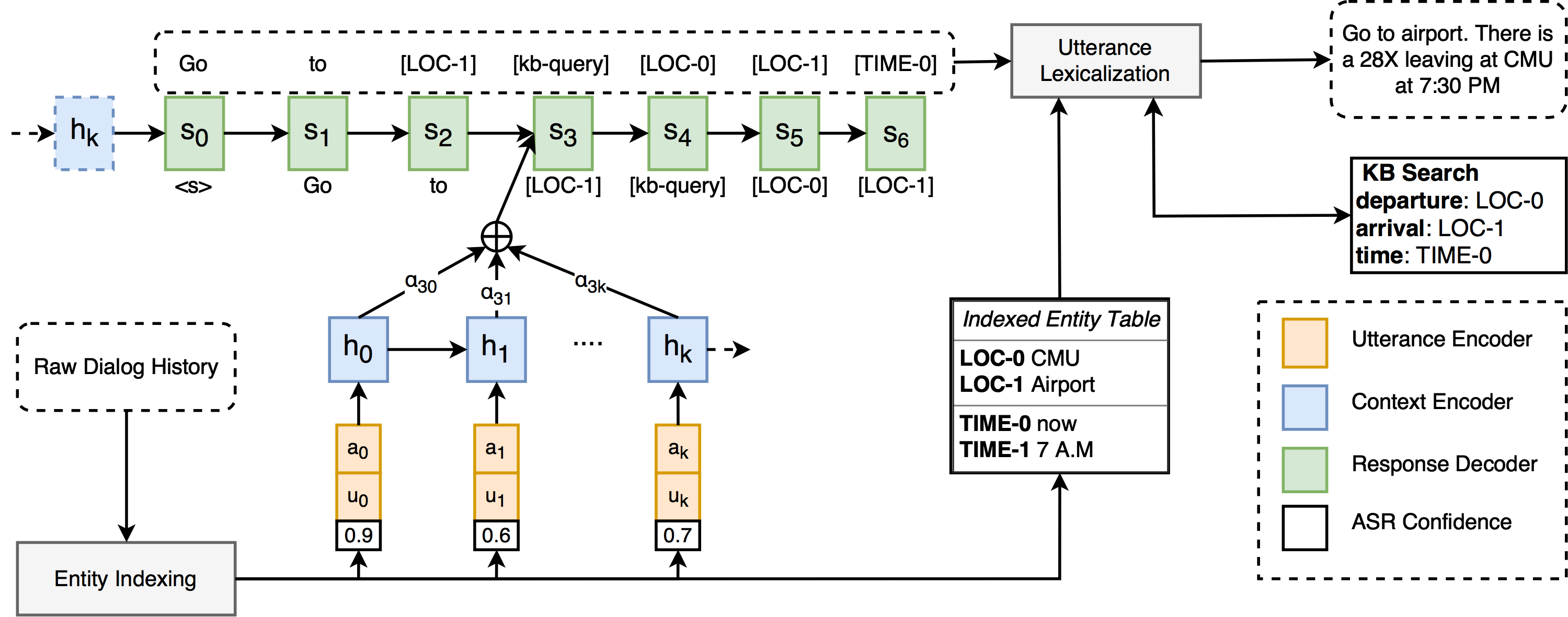}
    \caption{The proposed pipeline for task-oriented dialog systems.}
    \label{fig:model}
\end{figure*}

Our proposed framework consists of three steps as shown in Figure~\ref{fig:model}: a) entity indexing (EI), b) slot-value independent encoder-decoder (SiED), c) system utterance lexicalization (UL). The intuition is to leverage domain-general named entity recognition (NER)~\cite{tjong2003introduction} techniques to extract salient entities in the raw dialog history and convert the lexical values of the entities into entity indexes. The encoder-decoder model is then trained to focus solely on reasoning over the entity indexes in a dialog history and to make decisions about the next utterance to produce (including KB query). In this way, the model can be unaffected by the inclusion of new entities and new KB, while maintaining its domain-general input/output interface for easy extension to new types of conversation skills. Lastly, the output from the decoder networks are lexicalized by replacing the entity indexes and special KB tokens with natural language. The following sections explain each step in detail. 

\subsection{Entity Indexing and Utterance Lexicalization}
\begin{figure}
    \centering
    \includegraphics[width=0.47\textwidth]{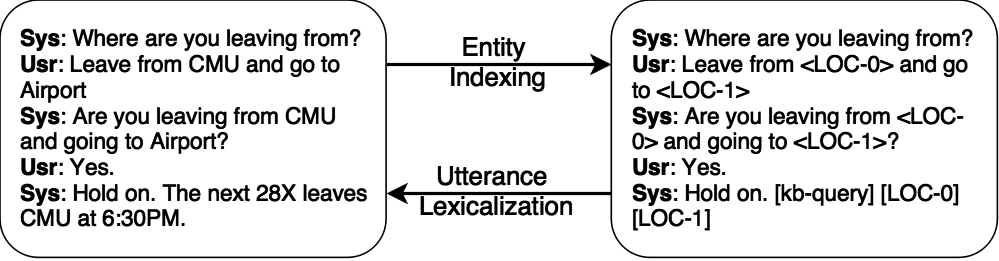}
    \caption{An example of entity indexing and utterance lexicalization.}
    \label{fig:example}
\end{figure}

\textbf{Entity Indexing} EI has two parts. First, the EI utilizes an existing domain-general NER to extract entities from both the user and system utterances. Note that the entity here is assumed to be a super-set of the slots in the domain. For example, for a flight-booking system, the system may contain two slots: [from-LOCATION] and [to-LOCATION] for the departure and arrival city, respectively. However, EI only extracts every mention of [LOCATION] in the utterances and leaves the task of distinguishing between departure and arrival to the encoder-decoder model. Furthermore, this step replaces each KB search result with its search query (e.g. the weather is cloudy $\rightarrow$ [kb-search]-[DATETIME-0]). The second step of EI involves constructing a \textit{indexed entity table}. Each entity is indexed by its order of occurrence in the conversation. Figure~\ref{fig:example} shows an example in which there are two [LOCATION] mentions. 

\textbf{Properties of Entity Indexing}
In this section, several properties of EI and their assumptions are addressed. First, each entity is indexed uniquely by its entity type and index. Note that the index is not associated with the entity value, but rather solely by the order of appearance in the dialog. Despite the actual words being hidden, a human can still easily predict which entity the system should confirm or search for in the KB based on logical reasoning. Therefore, that the EI not only alleviates the OOV problem of deploying the encoder-decoder model in the real world, but also forces the encoder-decoder model's focus on learning the reasoning process of task-oriented dialogs instead of leveraging too much information from the language modeling.   

Moreover, most slot-filling SDSs, apart from informing the concepts from KBs, usually do not introduce novel entities to users. Instead, systems mostly corroborate the entities introduced by the users. With this assumption, every entity mention in the system utterances can always be found in the users' utterances in the dialog history, and therefore can also be found in the indexed entity table. This property reduces the grounding behavior of the conventional task-oriented dialog manager into selecting an entity from the indexed entity table and confirming it with the user.

\textbf{Utterance Lexicalization} is the reverse of EI. Since EI is a deterministic process, its effect can always be reversed by finding the corresponding entity in the indexed entity table and replacing the index with its word. For KB search, a simple string matching algorithm can search for the special [kb-search] token and take the following generated entities as the argument to the KB. Then the actual KB results can replace the original KB query. Figure~\ref{fig:example} shows an example of utterance lexicalization.

\subsection{Encoder-Decoder Models}
The encoder-decoder model can then read in the EI-processed dialog history and predict the system's next utterance in EI format. Specifically, a dialog history of $k$ turns is represented by $[(a_0, u_0, c_0), ... (a_{k-1}, u_{k-1}, c_{k-1})]$, in which $a_i$, $u_i$ and $c_i$ are, respectively, the system, user utterance and ASR confidence score at turn $i$. Each utterance in the dialog history is encoded into fixed-size vectors using Convolutional Neural Networks (CNNs) proposed in~\cite{kim2014convolutional}. Specifically, each word in an utterance $x$ is mapped to its word embedding, so that an utterance is represented as a matrix $R \in R^{|x| \times D}$, in which $D$ is the size of the word embedding. Then $L$ filters of size 1,2,3 conduct convolutions on $R$ to obtain a feature map, $c$, of n-gram features in window size 1,2,3. Then $c$ is passed through a nonlinear ReLu~\cite{glorot2011deep} layer, followed by a max-pooling layer to obtain a compact summary of salient n-gram features, i.e.
$e^t(x) = \text{maxpool}(\text{ReLu}(c + b))$. Using CNNs to capture word-order information is crucial, because the encoder-decoder has to be able to distinguish between fine-grained differences between entities. For example, a simple bag-of-word embedding approach will fail to distinguish between the two location entities in ``leave from [LOCATION-0] and go to [LOCATION-1]'', while a CNN encoder can capture the context information of these two entities.

After obtaining utterance embedding, a turn-level dialog history encoder network similar to the one proposed in~\cite{zhao2016towards} is used. Turn embedding is a simple concatenation of system, user utterance embedding and the confidence score $t=[e^u(a_i); e^u(u_i); c_i]$. Then an Long Short-Term Memory (LSTM)~\cite{hochreiter1997long} network reads the sequence turn embeddings in the dialog history via recursive state update $s_{i+1} = \text{LSTM}(t_{i+1}, h_i)$, in which $h_i$ is the output of the LSTM hidden state. 

\textbf{Decoding with/without Attention} A vanilla decoder takes in the last hidden state of the encoder as its initial state and decodes the next system utterance word by word as shown in~\cite{sutskever2014sequence}. This assumes that the fixed-size hidden state is expressive enough to encode all important information about the history of a dialog. However, this assumption may often be violated for a task that has long-dependency or complex reasoning of the entire source sequence. An attention mechanism proposed~\cite{bahdanau2014neural} in the machine translation community has helped encoder-decoder models improve state-of-art performance in various tasks~\cite{bahdanau2014neural,xu2015show}. Attention allows the decoder to look over every hidden state in the encoder and dynamically decide the importance of each hidden state at each decoding step, which significantly improves the model's ability to handle long-term dependency. We experiment decoders both with and without attention. Attention is computed similarly multiplicative attention described in~\cite{luong2015effective}. We denote the hidden state of the decoder at time step $j$ by $s_j$, and the hidden state outputs of the encoder at turn $i$ by $h_i$. We then predict the next word by
\vspace{-0.5cm}
\begin{align}
    a_{ji} &= \text{softmax}(h_i^{T} W_a s_j + b_a) \label{eq:attn}\\
    c_j &= \sum_i a_{ji} h_i \\
    \widetilde{s_j} &= tanh(W_s 
    \begin{bmatrix}
        s_j \\
        c_j
    \end{bmatrix}) \\
    p(w_j|s_j, c_j) &= \text{softmax}(W_o \widetilde{s_j})
\end{align}
The decoder next state is updated by $s_{j+1} = \text{LSTM}(s_j, e(w_{j+1}), \widetilde{s_j})$.

\subsection{Leveraging Chat Data to Improve OOD Recovery}
Past work has shown that simple supervised learning is usually inadequate for learning robust sequential decision-making policy~\cite{williams2003using,ross2011reduction}. This is because the model is only exposed to the expert demonstration, but not to examples of how to recover from its own mistakes or users' OOD requests. We present a simple yet effective technique that leverages the extensibility of the encoder-decoder model in order to obtain a more robust policy in the setting of supervised learning. Specifically, we artificially augment a task-oriented dialog dataset with chat data from an open-domain conversation corpus. This has been shown to be effective in improving the performance of task-oriented systems~\cite{yu2017learning}. Let the original dialog dataset with $N$ dialogs be $\mathcal{D} = [d_0 ...,d_n,... d_N]$, where $d_n$ is a multi-turn task-oriented dialog of $|d_n|$ turns. Furthermore, we assume we have access to a chat dataset $\mathcal{D}_c = [(q_0, r_0),...(q_m,r_m),...(q_M, r_M)]$, where $q_m$, $r_m$ are common adjacency pairs that appear in chats, (e.g. $q=$ hello, $r=$ hi, how are you). Then we can create a new dataset $\mathcal{D}^*$ by repeating the following process a certain number of times:
\begin{enumerate}
    \item Randomly sample dialog $d_n$ from $\mathcal{D}$
    \item Randomly sample turn $t_i=[a_i, u_i]$ from $d_n$ 
    \item Randomly sample an adjacency pair $(q_m, r_m)$ from $\mathcal{D}_c$
    \item Replace the user utterance of $t_i$ by $q_m$ so that $t_i=[a_i, q_m]$
    \item Insert a new turn after $t_i$, i.e. $t_{i+1} = [r_m+e_{i+1}, u_i]$
\end{enumerate}
\begin{figure}[ht]
    \centering
    \includegraphics[width=0.47\textwidth]{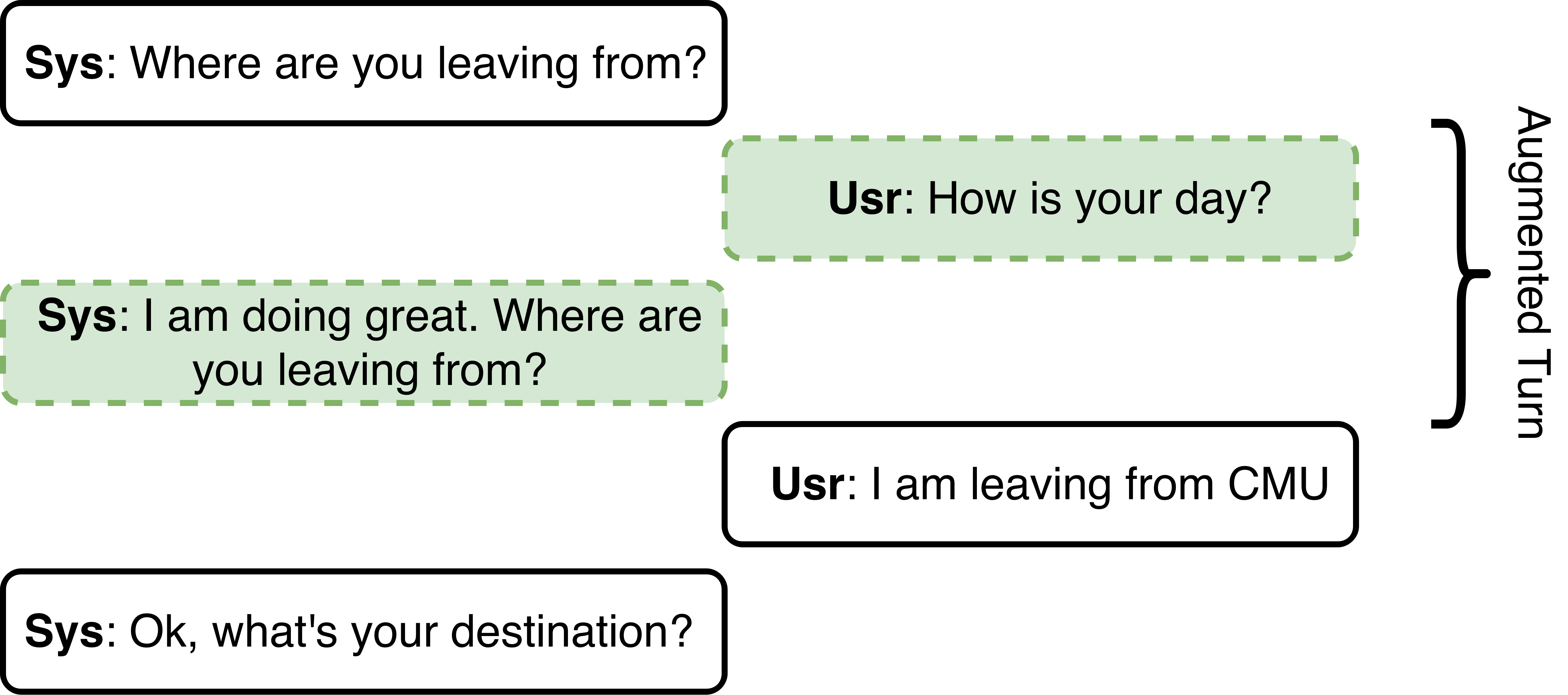}
    \caption{Illustration of data augmentation. The turn in the dashed line is inserted in the original dialog.}
    \label{fig:ood}
\end{figure}
In Step 5, $e_i$ is an error handling system utterance after the system answers the user's OOD request, $q_m$. In this study, we experimented with a simple case where $e_{i+1} = a_i$ so that the system should repeat its previous prompt after responding to $q_m$ via $r_m$. Figure~\ref{fig:ood} shows an example of an augmented turn. Eventually, we train the model on the union of the two datasets $\mathcal{D}^+ = \mathcal{D} \cup \mathcal{D}^*$

\textbf{Discussion}: There are several reasons that the above data augmentation process is appealing. First, the model effectively learns an OOD recovery strategy from $\mathcal{D}^*$, i.e. it first gives chatting answers to users' OOD requests and then tries to pull users back to the main-task conversation. Second, chat data usually has a larger vocabulary and more diverse natural language expressions, which can reduce the chance of OOVs and enable the model to learn more robust word embeddings and language models.  

\section{Experiment Setup}
\label{sec:setup}
\subsection{Dataset and Domain}
The CMU Let's Go Bus Information System~\cite{raux2005let} is a task-oriented spoken dialog system that contains bus information. We combined the train1a and train1b datasets from DSTC 1~\cite{williams2013dialog}, which contain 2608 total dialogs. The average dialog length is 9.07 turns. The dialogs were randomly splitted into 85/5/10 proportions for train/dev/test data. The data was noisy since the dialogs were collected from real users via telephone lines. Furthermore, this version of Let's Go used an in-house database containing the Port Authority bus schedule. In the current version, that database was replaced with the Google Directions API, which both reduces the human burden of maintaining a database and opens the possibility of extending Let's Go to cities other than Pittsburgh. Connecting to Google Directions API involves a POST call to their URL, with our given access key as well as the parameters needed: departure place, arrival place and departure time, and the travel mode, which we always set as TRANSIT to obtain relevant bus routes. There are 14 distinct dialog acts available to the system, and each system utterance contains one or more dialog acts. Lastly, the system vocabulary size is 1311 and the user vocabulary size is 1232. After the EI process, the sizes become 214 and 936, respectively.  

For chat data, we use a publicly available chat corpus used in~\cite{yu2015ticktock}\footnote{github.com/echoyuzhou/ticktock\_text\_api}. In total, there are 3793 chatting adjacency pairs. We control the number of data injections to 30\% of the number of turns in the original DTSC dataset, which leads to a user vocabulary size of 3537 and system vocabulary size of 4047.

\subsection{Training Details}
For all experiments, the word embedding size was 100. The sizes of the LSTM hidden states for both the encoder and decoder were 500 with 1 layer. The attention context size was also 500. We tied the CNN weights for the encoding system and user utterances. Each CNN has 3 filter windows, 1, 2, and 3, with 100 feature maps each. We trained the model end-to-end using Adam~\cite{kingma2014adam}, with a learning rate of 1e-3 and a batch size of 40. To combat overfitting, we apply dropout~\cite{zaremba2014recurrent} to the LSTM layer outputs and the CNN outputs after the maxpooling layer, with a dropout rate of 40\%.

\section{Experiments Results}
\label{sec:exp}
This approach was assessed both offline and online evaluations. The offline evaluation contains standard metrics to test open-domain encoder-decoder dialog models~\cite{li2015diversity,serban2015building}. System performance was assessed from three perspectives that are essential for task-oriented systems: dialog acts, slot-values, and KB query. The online evaluation is composed of objective task success rate, the number of turns, and subjective satisfaction with human users. 
\subsection{Offline Evaluation}

\textbf{Dialog Acts (DA):} Each system utterance is made up of one or more dialog acts, e.g. ``leaving at [TIME-0], where do you want to go?'' $\rightarrow$ [implicit-confirm, request(arrival place)]. To evaluate whether a generated utterance has the same dialog acts as the ground truth, we trained a multi-label dialog tagger using one-vs-rest Support Vector Machines (SVM)~\cite{tsoumakas2006multi}, with bag-of-bigram features for each dialog act label. Since the natural language generation module in Let's Go is handcrafted, the dialog act tagger achieved 99.4\% average label accuracy on a held-out dataset. We used this dialog act tagger to tag both the ground truth and the generated responses. Then we computed the micro-average precision, recall, and the F-score.

\textbf{Slots:} This metric measures the model's performance in generating the correct slot-values. The slot-values mostly occur in grounding utterances (e.g. explicit/implicit confirm) and KB queries. We compute precision, recall, and F-score.

\textbf{KB Queries:} Although the slots metric already covers the KB queries, here the precision/recall/F-score of system utterances that contain KB queries are also explicitly measured, due to their importance. Specifically, this action measures whether the system is able to generate the special [kb-query] symbol to initiate a KB query, as well as how accurate the corresponding KB query arguments are.

\textbf{BLEU}~\cite{papineni2002bleu}: compares the n-gram precision with length penalty, and has been a popular score used to evaluate the performance of natural language generation~\cite{wen2015semantically} and open-domain dialog models~\cite{li2016deep}. Corpus-level BLEU-4 is reported. 

\begin{table}[!ht]
\centering
\begin{tabular}{p{1.2cm}|p{1cm}p{1cm}p{1cm}p{1cm}} \hline
Metrics       & Vanilla        & EI             & EI +Attn        & EI+Attn +Chat   \\ \hline
DA (p/r/f1)   & 83.5 77.9 80.5 & 79.7 80.1 80.0 & 80.0 83.1 81.5 & 81.8 83.5 82.7 \\ \hline
Slot (p/r/f1) & 42.0 30.3 35.2 & 60.6 63.6 62.1 & 63.7 64.7 64.2 & 64.6 69.1 66.8 \\ \hline
KB (p/r/f1)   & N/A            & 48.9 55.3 51.9 & 55.4 70.8 62.2 & 58.2 71.9 64.4 \\ \hline
BLEU          & 36.9  & 54.6           & 59.3           & 60.5 \\\hline
\end{tabular}
    \caption{Performance of each model on automatic measures.}
    \label{tbl:results}
\end{table}

Four systems were compared: the basic encoder-decoder models without EI (vanilla), the basic model with EI pre-processing (EI), the model with attentional decoder (EI+Attn) and the model trained on the dataset augmented with chatting data (EI+Attn+Chat). The comparison was carried out on exactly the same held-out test dataset that contains 261 dialogs. Table~\ref{tbl:results} shows the results. It can be seen that all four models achieve similar performance on the dialog act metrics, even the vanilla model. This confirms the capacity of encoder-decoders models to learn the ``shape'' of a conversation, since they have achieved impressive results in more challenging settings, e.g. modeling open-domain conversations. Furthermore, since the DSTC1 data was collected over several months, there were minor updates made to the dialog manager. Therefore, there are inherent ambiguities in the data (the dialog manager may take different actions in the same situation). We conjecture that $\sim$80\% is near the upper limit of our data in modeling the system's next dialog act given the dialog history. 

On the other hand, these proposed methods significantly improved the metrics related to slots and KB queries. The inclusion of EI alone was able to improve the F-score of slots by a relative 76\%, which confirms that EI is crucial in developing slot-value independent encoder-decoder models for modeling task-oriented dialogs. Likewise, the inclusion of attention further improved the prediction of slots in system utterances. Adding attention also improved the performance of predicting KB queries, more so than the overall slot accuracy. This is expected, since KB queries are usually issued near the end of a conversation, which requires global reasoning over the entire dialog history. The use of attention allows the decoder to look over the history and make better decisions rather than simply depending on the context summary in the last hidden layer of the encoder. Because of the good performance achieved by the models with the attentional decoder, the attention weights in Equation~\ref{eq:attn} at every step of the decoding process in two example dialogs from test data are visualized. For both figures, the vertical axes show the dialog history flowing from the top to the bottom. Each row is a turn in the format of (system utterance \# user utterance). The top horizontal axis shows the predicted next system utterance. The darkness of a bar indicates the value of the attention calculated in Equation~\ref{eq:attn}.

The first example shows attention for grounding the new entity [LOCATION-1] in the previous turn. The attention weights become focus on the previous turn when predicting [LOCATION-1] in the implicit confirm action. The second dialog example shows a more challenging situation, in which the model is predicting a KB query. We can see that the attention weights when generating each input argument of the KB query clearly focus on the specific mention in the dialog history. The visualization confirms the effectiveness of the attention mechanism in dealing with long-term dependency at discourse level.
\begin{figure}[ht]
    \centering
    \includegraphics[width=0.48\textwidth]{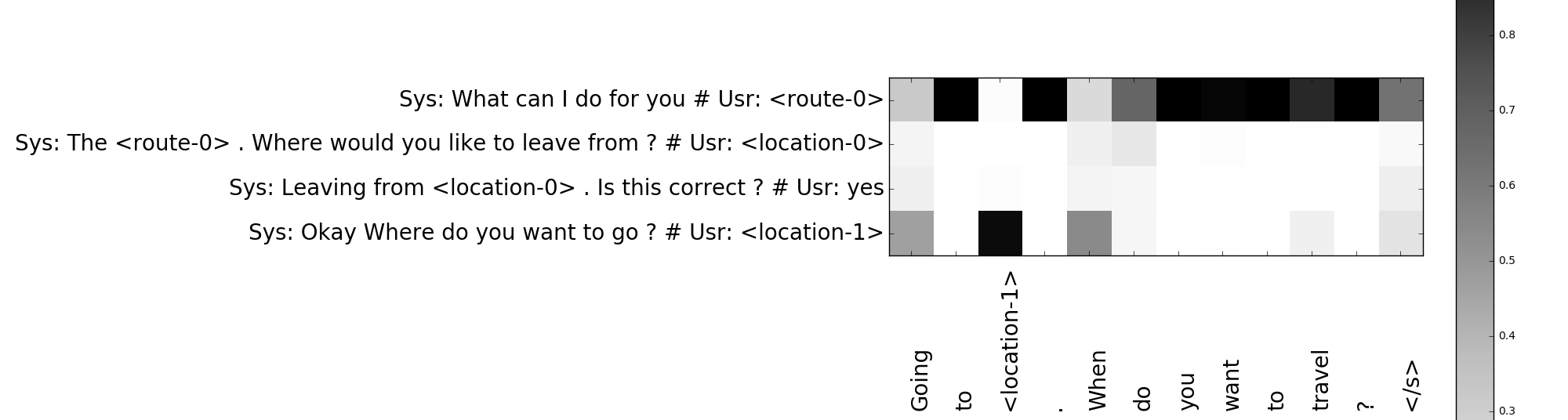}
    \includegraphics[width=0.48\textwidth]{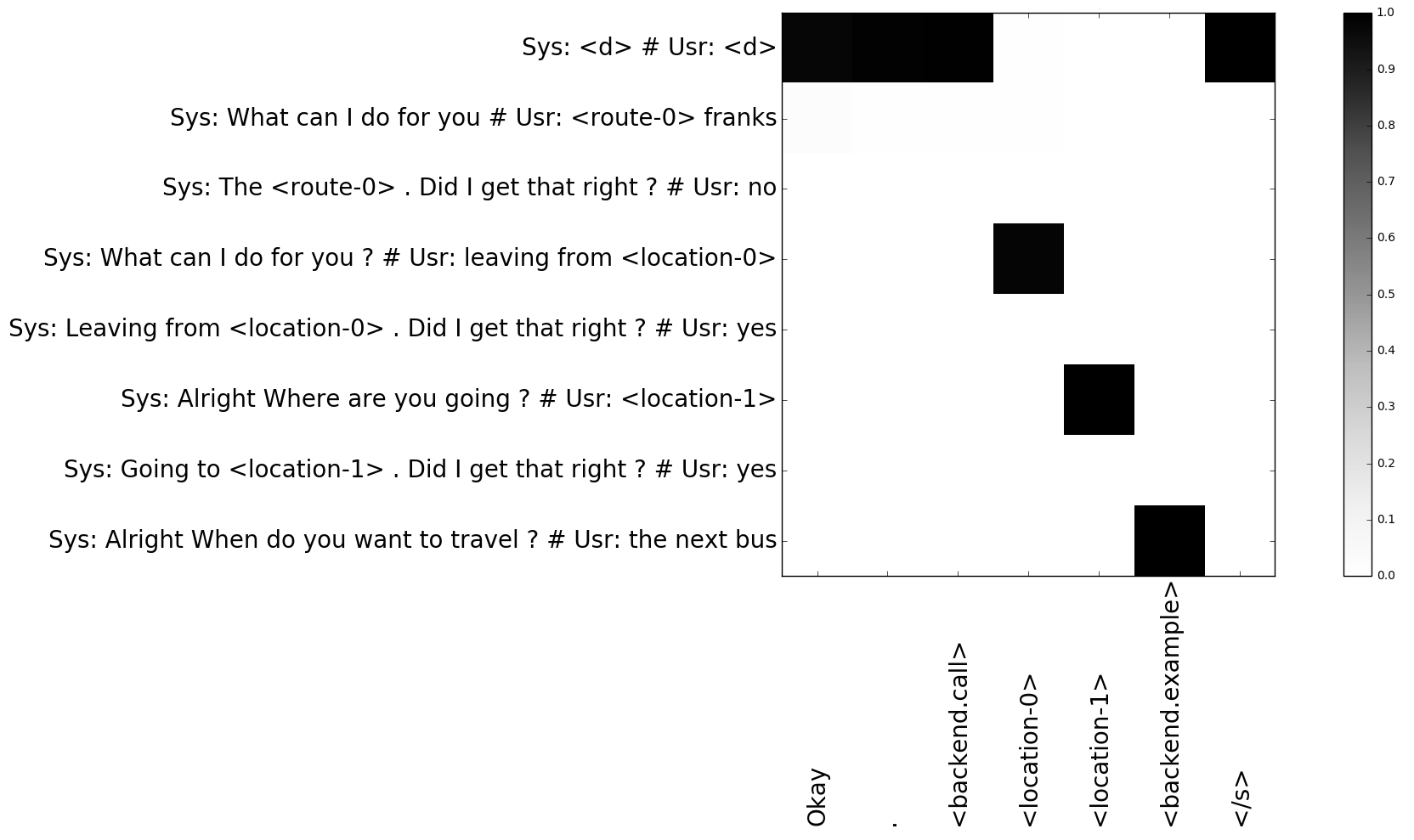}
    \caption{Visualization of attention weights when generating implicit confirm (top) and KB query (bottom).}
    \label{fig:attn}
\end{figure}

Surprisingly, the model trained on the data augmented with chat achieved slightly better slot accuracy performance, even though the augmented data is not directly related to task-oriented dialogs. Furthermore, the model trained on chat-augmented data achieved better scores for the KB query metrics. Several reasons may explain this improvement: 1) since chat data exposes the model to a significantly larger vocabulary, the resulting model is more robust to words that it had not seen in the original task-oriented-only training data, and 2) the augmented dialog turn can be seen as noise in the dialog history, which adds extra regularization to the model and enables the model to learn more robust long-term reasoning mechanisms. 

\subsection{Human Evaluation}
Although the model achieves good performance in offline evaluation, this may not carray over to  real user dialogs, where the system must simultaneously deal with several challenges, such as automatic speech recognition (ASR) errors, OOD requests, etc. Therefore, a real user study was conducted to evaluate the performance of the proposed systems in the real world. Due to the limited number of real users, only two best performing system were compared, EI+Attn and EI+Attn+Chat. Users were able to talk to a web interface to the dialog systems via speech. Google Chrome Speech API~\footnote{www.google.com/intl/en/chrome/demos/speech.html} served as the ASR and text-to-speech (TTS) modules. Turn-taking was done via the built-in Chrome voice activity detection (VAD) plus a finite state machine-based end-of-turn detector~\cite{zhao2015incremental}. Lastly, a hybrid named entity recognizer (NER) was trained using Conditional Random Field (CRF)~\cite{mccallum2003early} and rules to extract 4 types of entities (location, hour, minute, pm/am) for the EI process. 

The experiment setup is as follows: when a user logs into the website, the system prompts the user with a goal, which is a randomly chosen combination of departure place, arrival place and time (\textit{e.g. leave from CMU and go to the airport at 10:30 AM}). The system also instructs the user to say goodbye if the he/she thinks the goal is achieved or wants to give up. The user begins a conversation with one of the two evaluated systems, with a 50/50 chance of choosing either system (not visible to the user). After the user's session is finished, the system asks the him/her to give two scores between 1 and 5 for correctness and naturalness of the system respectively. The subjects in this study consist of undergraduate and graduate students. However, many subjects did not follow the prompted goal, but rather asked about bus routes of their own. Therefore, the dialog was manually labeled for dialog success. A dialog is successful if and only if the systems give at least one bus schedule that matches with all three slots expressed by the users. 
\begin{table}[ht]
    \centering
    \begin{tabular}{p{0.16\textwidth}|p{0.12\textwidth}p{0.12\textwidth}} \hline 
    Metrics          & EI+Attn & EI+Attn +Chat \\ \hline
    \# of Dialog    & 75               & 74 \\ 
    Slot Precision   & 73.3\%           & 71.8\% \\
    KB Precision     & 88.6\%           & 93.7\% \\ 
    Success Rate     & 73.3\%           & 77.0\% \\ 
    Avg Turns        & 4.88             & 4.91 \\ \hline
    Avg Correctness  & 3.45 (1.32)      & 3.22 (1.40)\\
    Avg Naturalness  & 3.46 (1.41)      & 3.53 (1.34)\\ \hline
    \end{tabular}
    \caption{Performance of each model on automatic measures. The standard deviations of subjective scores are in parentheses.}
    \label{tbl:usr}
\end{table}
Table~\ref{tbl:usr} shows the results. Overall, our systems achieved reasonable performance in terms of dialog success rate. The EI+Attn+Chat model achieves slightly higher success and subjective naturalness metrics (although the difference between EI+Attn+Chat and EI+Attn was not statistically significant due to the limited number of subjects). The precision of grounding the correct slots and predicting the correct KB query was also manually labelled. EI+Attn model performs slightly better than the EI+Attn+Chat model in slot precision, while the latter model performs significantly better in KB query precision. In addition, EI+Attn+Chat leads to slightly longer dialogs because sometimes it generates chatting utterances with users when it cannot understand users' utterances. 

At last, we investigated the log files and identified the following major types of sources of dialog failure:
\textbf{RNN Decoder Invalid Output:} Occasionally, the RNN decoder outputs system utterances as ``Okay going to [LOCATION-2]. Did I get that right?'', in which [LOCATION-2] cannot be found in the indexed entity table. Such invalid output confuses users. This occurred in 149 of the dialogs, where 4.1\% of system utterances contain invalid symbols. \textbf{Imitation of Suboptimal Dialog Policy:} Since our models are only trained to imitate the suboptimal hand-crafted dialog policy, their limitations show when the original dialog manager cannot handle the situation, such as failing to understand slots that appeared in compound utterances. Future plans involves improving the models to perform better than the suboptimal teacher policy.

\section{Conclusions}
\label{sec:conclusion}
In conclusion, this paper discusses constructing task-oriented dialog systems using generative encoder decoder models. EI is effective in solving both the OOV entity and KB query challenges for encoder-decoder-based task-oriented SDSs. Additionally, the novel data augmentation technique of interleaving task-oriented dialog corpus with chat data led to better model performance in both online and offline evaluation. Future work includes developing more advanced encoder-decoder models that to better deal with long-term dialog history and complex reasoning challenges than current models do. Furthermore, inspired by the success of mixing chatting with slot-filling dialogs, we will take full advantage of the extensibility of encoder-decoder models by investigating how to make systems that are able to interleave various conversational tasks, e.g. different domains, chatting or task-oriented, which in turn can create a more versatile conversational agent. 


\newpage
\bibliography{acl2017}
\bibliographystyle{acl_natbib}

\appendix\end{document}